\documentclass{article}
\usepackage{spconf,amsmath,graphicx}
\usepackage{booktabs}
\usepackage{pifont}
\usepackage{multirow}
\usepackage{amsthm}
\usepackage{color}
\usepackage{url}

\title{A FAST, PERFORMANT, SECURE DISTRIBUTED TRAINING FRAMEWORK FOR LLM}
\name{Wei Huang$^*$, Yinggui Wang$^*\dag$ \thanks{$^*$These authors contributed equally to this work} \thanks{$^\dag$  Corresponding anthor (wyinggui@gmail.com).}, Anda Cheng, Aihui Zhou, Chaofan Yu, Lei Wang}
\address{Ant Group, China}
\begin{document}
%
\maketitle
\begin{abstract}
The distributed (federated) LLM is an important method for co-training the domain-specific LLM using siloed data. However, maliciously stealing model parameters and data from the server or client side has become an urgent problem to be solved. In this paper, we propose a secure distributed LLM based on model slicing. In this case, we deploy the Trusted Execution Environment (TEE) on both the client and server side, and put the fine-tuned structure (LoRA or embedding of P-tuning v2) into the TEE. Then, secure communication is executed in the TEE and general environments through lightweight encryption. In order to further reduce the equipment cost as well as increase the model performance and accuracy, we propose a split fine-tuning scheme. In particular, we split the LLM by layers and place the latter layers in a server-side TEE (the client does not need a TEE). We then 
combine the proposed Sparsification Parameter Fine-tuning (SPF) with the LoRA part to improve the accuracy of the downstream task. Numerous experiments have shown that our method guarantees accuracy while maintaining security.
\end{abstract}
\vspace{-0.1cm}
\section{Introduction}
\label{sec:intro}

The representative Parameter Efficient Fine-Tuning (PEFT) \cite{peft} methods for the LLM
include Low-Rank Adaptation (LoRA) \cite{hu2022lora}, Prompt Tuning v2 (P-tuning v2) \cite{liu2021p}. 
The limited amount of domain-specific data available at each institution limits the accuracy of the model. As a result, distributed (federated) LLM \cite{kuang2023federatedscopellm,zhang2023towards} has become a direction of great interest. 
Given the efficient learning and memorization ability of the data by the LLM, it leads to some security issues in the training phase of the distributed LLM: 1) A malicious server will steal parameters and infer data from the model. 2) Malicious clients can also infer data from the rest of the clients through model parameters and intermediate embedding. Previous work considered server-side threats, but did not consider the leakage of parameters and data in the client at the same time \cite{mothukuri2021survey,wang2019beyond,8241854}. Based on the above considerations, how the distributed LLM can avoid leakage of model parameters and data 
has become an urgent challenge.

To overcome the above problems, we propose a fast, high-accuracy and secure distributed LLM based on model slicing. It includes various implementations such as LoRA, P-Tuning v2, and the split fine-tuning proposed in this paper. And security is ensured through One Time Pad (OTP) \cite{tramer2018slalom} and the TEE \cite{mckeen2016intel,tdx}. In particular, when training in a distributed manner, we only fine-tune the partial parameters. For different types
of TEE, i.e., {\bf SGX (small-memory and no GPU)} \cite{mckeen2016intel} and  {\bf Intel
TDX/SGX (large memory and no GPU)} \cite{tdx,mckeen2016intel}, we propose different schemes in Section 3.1 and 3.2. 
The main contributions of this paper are as follows. We propose a secure distributed training framework for the LLM based on model slicing to solve the leakage problem of model parameters and data on the server and client side. And we effectively combine the TEE and lightweight encryption to ensure security.  Then, we also propose a new split fine-tuning strategy. Experiments demonstrate that our method guarantees high efficiency and accuracy even with security.



\vspace{-0.1cm}
\section{RELATE WORK}
\label{sec:format}
\vspace{-0.2cm}
\subsection{Security Threats to Federated Learning}

Federated learning has been found to face a variety of security threats \cite{mothukuri2021survey}. One of the threats we consider is the malicious server attack. Wang et al \cite{wang2019beyond}, proposed a malicious server attack that utilizes an adversarial generative network of multi-task discriminators to restore typical data for each client. Phong et al \cite{8241854}. proposed an aggregation method for privacy preservation using Additively Homomorphic Encryption (AHE) to defend against malicious server attacks. Agarwal et al \cite{agarwal2018cpsgd}, considering the scenario where the client does not trust the server, proposed a distributed Stochastic Gradient Descent (SGD) by utilizing a binary term mechanism to perturb the upload gradient, which is communication efficient and satisfies Differential Privacy (DP). However, DP can only protect the data but not the parameter leakage, and can cause significant performance degradation. Wagh et al \cite{wagh2019securenn}, proposed a tripartite secure computing framework for neural network training. But MPC will greatly increase training time. 
\vspace{-0.7cm}
\subsection{Trusted Execution Environment (TEE)}

A Trusted Execution Environment (TEE) is an isolated hardware enclave that stores and processes sensitive data. Popular TEE implementations include Intel SGX \cite{mckeen2016intel}, AMD SEV \cite{kaplan2016amd}, Intel TDX \cite{tdx}, and TrustZone \cite{alves2004trustzone}. In this paper, we follow prior work and deem TEE as a secure area on a potential adversary host device (including GPUs) \cite{hou2021model,shen2022model}. It means the data, code, and the whole computation process inside TEEs are secure. 

\section{METHOD}
\label{sec:pagestyle}
\subsection{Method1: Small memory (Consumer-grade) TEE-shielded LLM Partition}
\label{3.1}

In a distributed scenario, suppose there are $K$ clients. Let $D_{k}=\left\{ \left( x^{m},y^{m}\right)  \right\}^{N}_{m=1} k=1,...,K $ denote the training set of the $k$th client. 
We use $W_{k}$ to refer to the parameters that the $k$th client needs to update. The federated training of Method1 can be seen in Figure 1(a), where only the parameters of LoRA and the embedding of P-Tuning v2  are communicated between the server and clients. We use OTP to encrypt the communicated gradient and decrypt it in TEE. Then at round $t$, the global model parameters $W^{'}$ can be expressed as $W^{'}=\sum^{K}_{k=1} \frac{n_{k}}{n} W_{k}$, where $n=\left| D\right|  =\sum^{K}_{k=1} n_{k}$ is the total number of the global joint data. To address the leakage of model parameters and data on the server side, we deploy TEE(SGX) on the server side and do parameter aggregation in it. In training, we fix the parameters except $W_{k}$ to learn the model parameters $\theta$ with the following objective function:
\begin{equation}
\begin{aligned}
	\mathop{\arg\min}\limits_{\theta} \left[ L\left( \theta \right)  =\sum^{K}_{k=1} \frac{n_{k}}{n} L_{k}\left( \theta \right)\right]
\end{aligned}
\end{equation}%
where $L$ is the loss function (Cross-Entropy).

\begin{figure*}[htb]
\begin{minipage}[b]{0.48\linewidth}
  \centering
  \centerline{\includegraphics[width=0.8\linewidth]{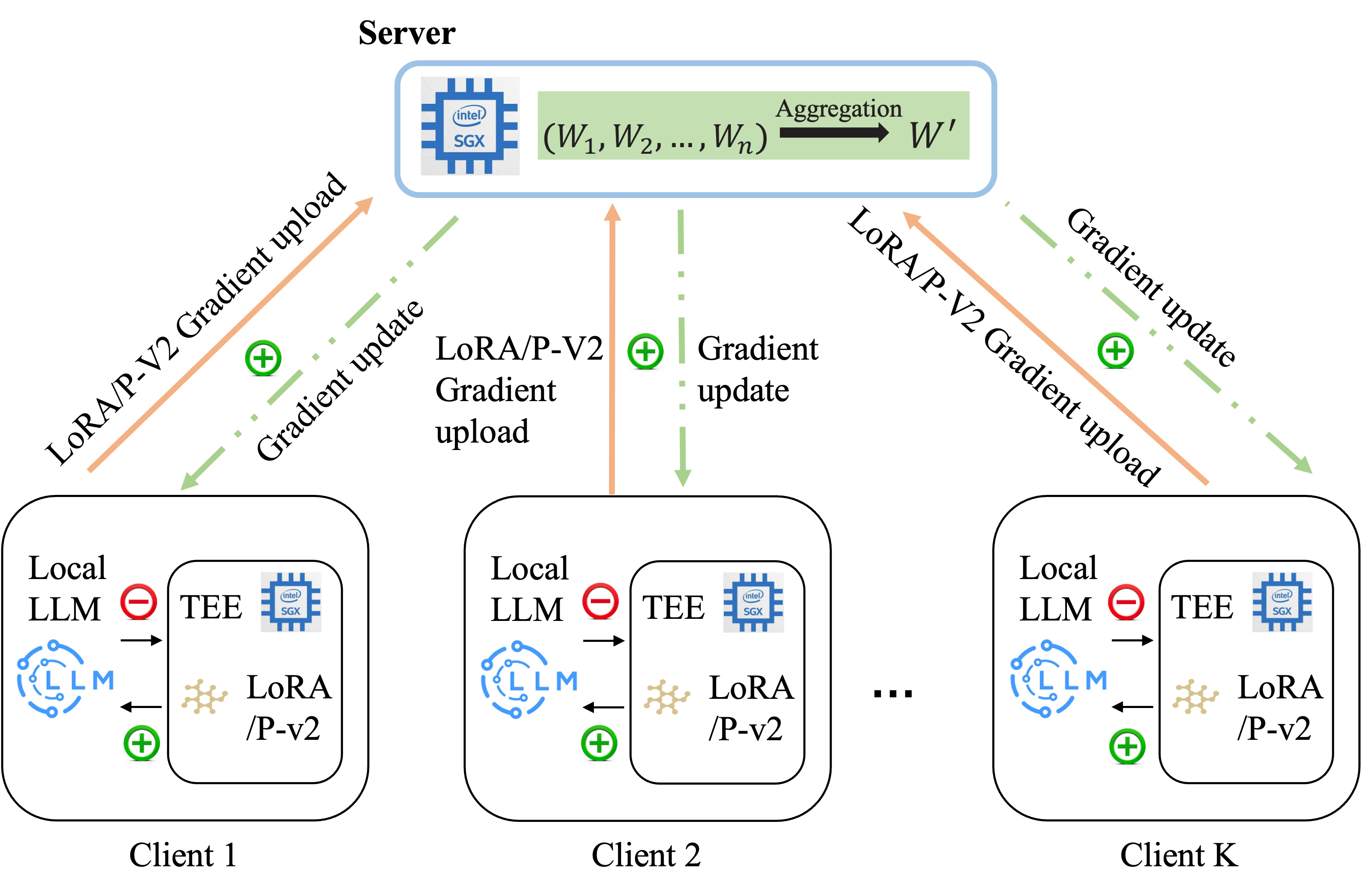}}
\vspace{-0.2cm}
  \centerline{(a) Distributed LLM training.}\medskip
\end{minipage}
\hfill
\begin{minipage}[b]{0.48\linewidth}
  \centering
  \centerline{\includegraphics[width=0.95\linewidth]{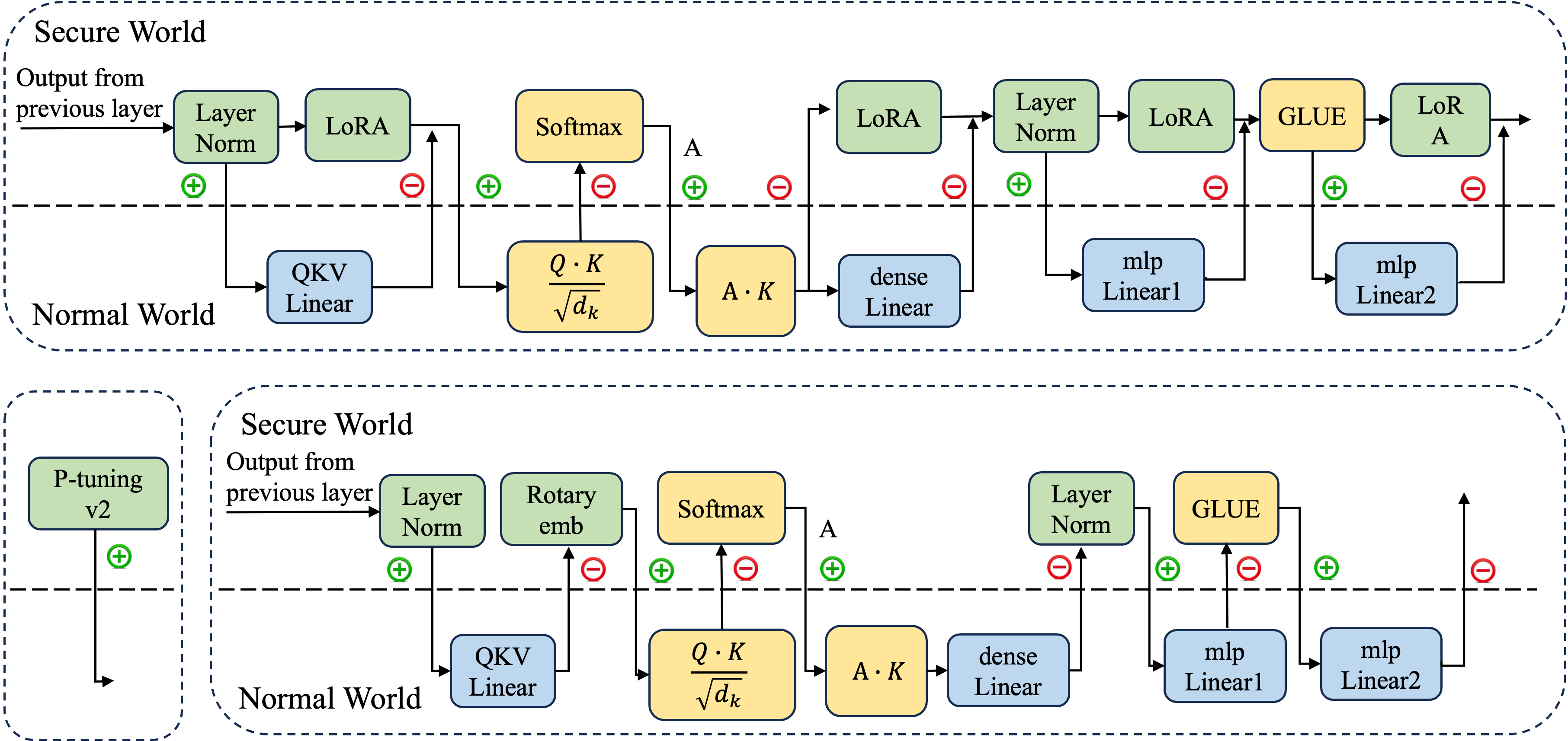}}
  \centerline{(b) Our TEE-shielded LLM Partition for Method1.}\medskip
\end{minipage}
\vspace{-0.3cm}
\caption{Block diagram of the distributed LLM based on the model partition for Method1, where the green plus sign indicates encryption and the red minus sign indicates decryption.}
\label{fig:res}
\end{figure*}

\label{sec:typestyle}

Note that the current client can maliciously infer the data of the rest clients from the parameters and intermediate embedding of the local model. To avoid the above problems, we deployed a TEE (SGX) on each client and put the model's LoRA and P-Tuning v2 embedding into this device. Since the TEE (SGX) has small memory and is cheap, each client can afford the expense of the device. However, the TEE needs to interact with external computing environments (e.g., on a general GPU). For the output embedding of the TEE carrying a large amount of user information, it also poses a large privacy risk without protecting it. We address the above problem using OTP to encrypt the features transmitted between the GPU and the TEE. Since OTP only supports encrypted computing for linear operations, in addition to the structure that needs fine-tuning, the non-linear layers of the model and the operations are shielded by the TEE, including the layer norm, activation function and softmax. Specific divisions can be seen in Figs. 1(a) and 1(b).

\textbf{Feature  Encryption.} For the LLM with a linear layer $h(.)$, we use $ E$ to be as the plaintext output of the TEE. Then we generate a random mask $ r$ and encrypt the features by the following formula: $E_{en}=E+r$,
where $E_{en}$ denotes the encrypted feature.

\textbf{Feature  Decryption.} The GPU receives $E_{en}$ , computes ${h(E}_{en})$ , and returns the result to the TEE. The decryption formula in the TEE is  $h\left( E\right)  =h\left( E{}_{en}\right)  -h\left(r\right).$
Since $Q \cdot K$ ($Q$ and $K$ represent specific linear layers in the attention module) involves matrix multiplication, we run this calculation on the GPU considering that it is more time-consuming for the TEE operations. The decryption formula for matrix multiplication is as follows:
\begin{align}
	\begin{gathered}Q_{E}\cdot K_{E}=Q_{E}\cdot K_{En}-Q_{E}\cdot r_{k} \\
    =Q_{En}\cdot K_{En}-Q_{En}\cdot r_{k}-r_{q}\cdot K_{En}+r_{q}\cdot r_{k}\end{gathered},
\end{align}%
where $Q_E$ and $K_E$ refer to the plaintext matrices of $Q$ and $K$, $Q_{En}$ and $K_{En}$ refer to the ciphertext matrices of $Q$ and $K$, and $r$ is the corresponding mask data.

Method1 slices the LLM, deploys the sensitive part of the structure in the TEE, and protects the transmission between the TEE and GPU through OTP to achieve the protection effect of the model parameters and data. 
\vspace{-0.2cm}
\subsection{Method2: Large memory TEE-shielded LLM Partition}
\vspace{-0.1cm}
Method1 solves the security problem, but its client generates a higher number of transmissions between the GPU and TEE during training. 
Secondly, during the process of encryption and decryption, there is a loss of numerical accuracy due to truncation errors, resulting in a slight degradation of accuracy. To alleviate the above problems, we propose a split fine-tuning method in Method2 and apply it to the distributed LLM. The model structure and fine-tuning methodology can be seen in Figure 2.

\begin{figure}[htb]
  \centering
  \centerline{\includegraphics[width=1\linewidth]{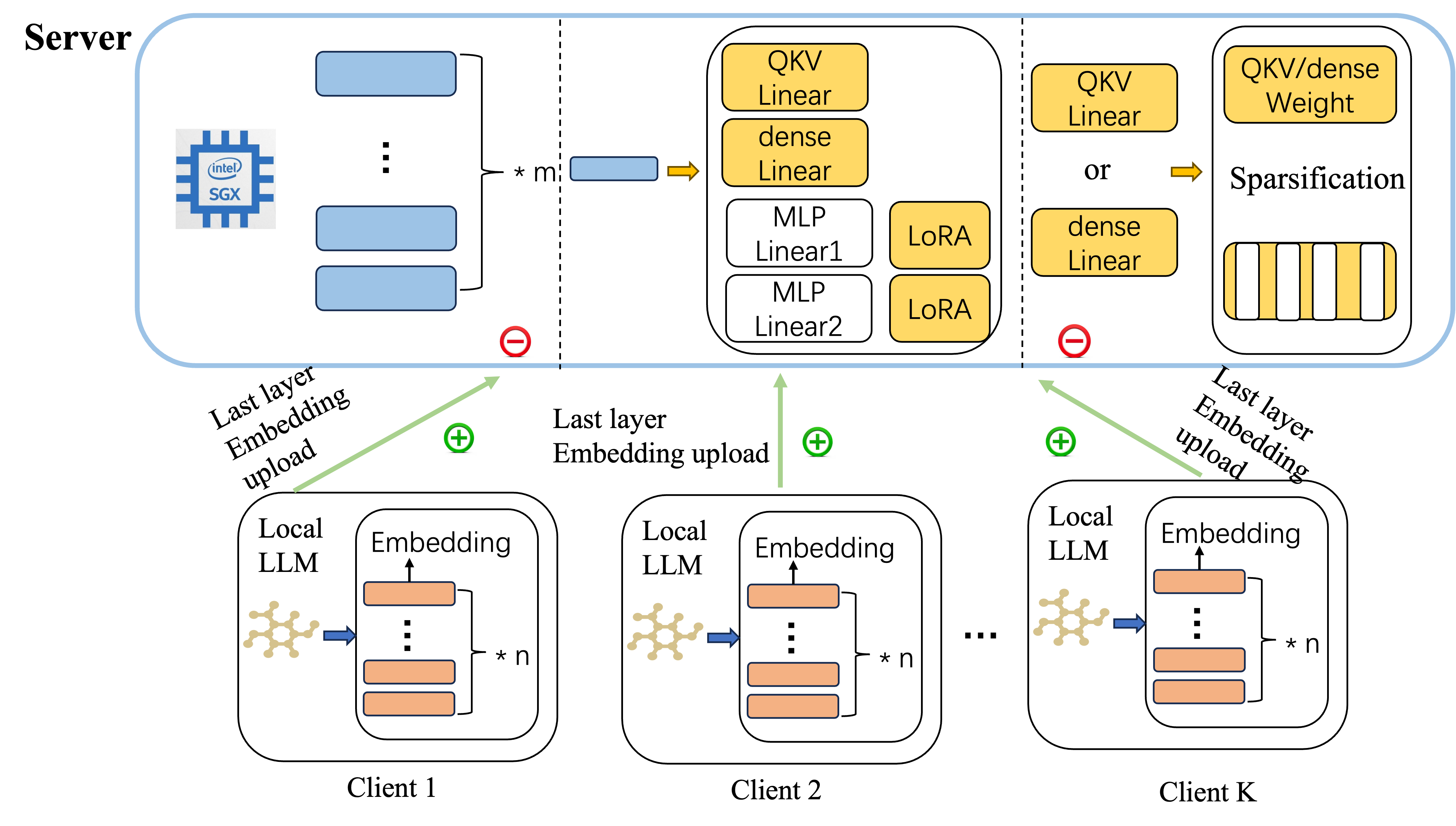}}
  \vspace{-0.3cm}
 \caption{The distributed LLM training for Method2.}
\label{fig:res1}
\end{figure}
{\bf Method2}. The model is first sliced by layer into two parts. Let $M_f$ refer to the first half of the model, $M_l$ the second half of the model, and $E_f$ the output features of the last layer of the first part. For each client, we deploy $M_f$. For the server, we deploy an TEE (Intel TDX) and put $M_l$ into that device. Since $M_l$ occupies a large amount of memory and the server side is used to provide external services with adequate resources, we use TEE (Intel TDX) as the device for this scheme. In distributed training, each client freezes the parameters of $M_f$ and saves $E_f$ generated for each piece of data. After each client collects the embedding of all the data, it is uploaded to the server's TEE using the OTP encryption. The server's TEE receives the data and does the decryption to obtain $E_f$, and then each $E_f$ is used as an input for $M_l$ to fine-tune the model. 

The strategy for fine-tuning is shown in Fig. 2. In particular, we fine-tune the QKV Linear and dense Linear for each part in $M_l$, for 
which we propose the SPF strategy based on the consideration that parameters with larger modulus values play a greater role in the model than parameters with smaller values. Specifically, we use $W_{all}$ to denote the weights of the linear layers and compute the L1 norm for each head on the second dimension of $W_{all}$. 
We select the heads by L1 norm value from large to small. After that, we recompose a certain proportion of these heads into the weights denoted as $W_{train}$ ($W_{freeze}$ denotes the weights of the remaining heads). During the fine-tuning process, we freeze $W_{freeze}$ to update $W_{train}$ only. We use  $X_{train}$ and $X_{freeze}$  to denote the inputs  of the linear layers. The forward formula of the SPF can be expressed as follows:
\begin{equation}
\begin{aligned}
	O =(X_{train} \cdot W_{train}^{T}+b_{train}) + (X_{freeze} \cdot W_{freeze}^{T} \\
    +b_{freeze}).
\end{aligned}
\end{equation}
The SPF method alleviates the problem of requiring a large number of fine-tuning parameters for fully fine-tuning the QKV linear and dense linear of $M_l$.
To further reduce the number of fine-tuning parameters, we choose to freeze the MLP parameters. To ensure performance, we fine-tune LoRA spliced to the MLP layer. 
\renewcommand{\tabcolsep}{2.0pt}
    \begin{table*}[htbp]
        \footnotesize
    	\begin{center}
    			{
    				\renewcommand{\arraystretch}{1.10}
    			\begin{tabular}{c|c|c|c|c|c|c|c}
                    \bottomrule[0.8pt]
    				 & &CHIP-CTC  &KUAKE-IR &KUAKE-QIC &KUAKE-QQR &KUAKE-QTR &Number of parameters (10 million) \\ \hline
                    \multirow{2}*{Method1}	&LoRA &0.822	&0.702	&0.815 &0.630 &0.483 &\textbf{1.5} \\
                    &P-Tuning v2 &0.816	&0.687	&0.811 &0.630 &0.457 &2.9 \\ \hline
                    Method2 &4 layers in TEE (25\%,50\%) &\textbf{0.858}	&\textbf{0.750}	&\textbf{0.845} &0.633 &\textbf{0.497} &8.5 \\ \hline
                    \multirow{2}*{FL-LLM}	&LoRA	&0.827	&0.730	&0.812 &0.633 &0.473 &\textbf{1.5} \\
                    &P-Tuning v2	 &0.818	&0.720	&0.816 &0.620 &0.453 &2.9 \\ \hline
                    \multirow{2}*{SWMT}	&LoRA	&0.825	&0.730	&0.811 &\textbf{0.635} &0.474 &\textbf{1.5} \\
                    &P-Tuning v2 &0.820	&0.724 &0.813 &0.622 &0.451 &2.9 \\ \toprule[0.8pt]
    			\end{tabular}
    		}
      \vspace{-0.2cm}
    		\caption{Comparison of the accuracy of different methods on datasets and the amount of parameters that need to be fine-tuned.}\label{table1}
      \vspace{-0.3cm}
    	\end{center}
    \end{table*}
    \begin{table*}[htbp]
        \footnotesize
    	\begin{center}
    			{
    				\renewcommand{\arraystretch}{1.10}
    			\begin{tabular}{c|c|c|c|c|c|c|c|c}
    				\bottomrule[0.8pt]
    				 & & The ratio &CHIP-CTC  &KUAKE-IR &KUAKE-QIC &KUAKE-QQR &KUAKE-QTR &Number of parameters (10 million) \\ \hline
                    \multirow{3}*{Method2}	&2 layers in TEE &100\%	, 100\%&0.858	&0.741	&0.839 &0.623 &0.503 &13.5 \\ 
                    &4 layers in TEE &100\% , 100\%&0.867	&0.733	&\textbf{0.858} &0.657 &0.540 &27.0 \\ 
                    &8 layers in TEE &100\% , 100\%&\textbf{0.884}	&\textbf{0.770}	&0.852 &\textbf{0.677}
                    &\textbf{0.587} &54.0 \\ \hline
                    \multirow{6}*{Method2} &4 layers in TEE &12.5\% , 100\%	&0.849	&0.723	&0.827 &0.633 &\textbf{0.543} &9.4 \\
                    &4 layers in TEE &25\%	, 100\%&0.862	&\textbf{0.753}	&0.842 &\textbf{0.650} &0.503 &11.8 \\ 
                    &4 layers in TEE &12.5\% , 50\%&0.836	&0.723	&0.824 &0.620 &0.490 &\textbf{6.0} \\ 
                    &4 layers in TEE &12.5\% , 62.5\%	&0.847	&0.727	&0.821 &0.627 &0.511 &6.9 \\ 
                    &4 layers in TEE &25\% , 50\%	&0.858	&0.750	&0.845 &0.633 &0.497 &8.5 \\ 
                    &4 layers in TEE &25\% , 62.5\%	&\textbf{0.865}	&0.743	&\textbf{0.848} &0.637 &0.510 &9.3 \\ 
                    \toprule[0.8pt]                  
    			\end{tabular}
    		}
      \vspace{-0.2cm}
    		\caption{Comparison of model accuracy and parameter quantities after LLM is divided into different layers using Method2. }
      \label{table2}
    	\end{center}
    \vspace{-0.5cm}
    \end{table*}

There is no interaction between each client, so there is no need to deploy a TEE. This scheme not only prevents the embedding transmitted from the client to the server from being stolen, but also the gradient does not need to be transmitted back to the client to avoid further information leakage. Since the $M_f$ parameter is frozen and does not have to be updated, each client can generate $E_f$ offline and upload it once. Since the updated model parameters are not available to each client at the end of distributed training, the scheme is served outward by the server to receive queries from users. 
\renewcommand{\tabcolsep}{2.0pt}
    \begin{table}[b]
    \vspace{-0.3cm}
        \footnotesize
    	\begin{center}
    			{
    				\renewcommand{\arraystretch}{1.10}
    			\begin{tabular}{c|c|c|c|c}
    				\bottomrule[0.8pt]
    				 &Method1 &Method2  &FL-LLM &SWMT \\ \hline
                    TEE in the client side 	&\ding{51}	&\ding{55}	&\ding{55}	&\ding{51} \\
                     TEE in the server side 	&\ding{51}	&\ding{51}	&\ding{55}	&\ding{51} \\ 
                    The type of TEE	&SGX	&TDX/SGX	&-	&TDX/SGX \\
                    Secure training	&\ding{51}	&\ding{51}	&\ding{55}	&\ding{51} \\ \toprule[0.8pt]         
    			\end{tabular}
    		}
      \vspace{-0.2cm}
    		\caption{Comparison of different schemes, where the memory of SGX is small and that of Intel TDX/SGX is large.}\label{table3}
    	\end{center}
    \end{table}
    
\section{EXPERIMENTAL EVALUATION}
\subsection{Datasets and Experimental Configurations}
We select five medical datasets as the industry data for this experiment, namely, CHIP-CTC, KUAKE-IR, KUAKE-QIC, KUAKE-QQR, and KUAKE-QTR ~\cite{zhang2021cblue,cblue}. We select ChatGLM-6B ~\cite{zeng2023glm-130b} as the base LLM. For the experimental setup, we set the number of clients to be 3, the number of communication rounds for distributed training to also be 10. And each dataset is divided into three equal parts, which are assigned to different clients. For LoRA, we set the rank to be 8 and lora\_dropout to be 0.1. For P-Tuning V2, we set the sequence length to be 128. For Method2, we set the location of the split to be the 24th layer (more details in Section 4.3). For the QKV linear and dense linear, we choose the ratios (12.5\%,25\%) and (50\%,62.5\%), respectively. As far as we know, we are the first to propose to prevent both malicious theft of model parameters and data by the server and malicious theft by the client during distributed training of the LLM, and we will compare it with the plaintext (without any security measures) Federal LLM (FL-LLM) and Shielding-Whole-Model by the TEE (SWMT). Table \ref{table3} shows the comparison between our two proposed schemes and the remaining schemes. 
\vspace{-0.2cm}
\subsection{Experimental Results of Method1 and Method2}
\vspace{-0.1cm}
Table \ref{table1} shows the accuracy of Method1 and Method2 for the five datasets and the number of parameters that need to be fine-tuned. From the table, we can see that the average accuracy of Method1 on the five datasets is slightly lower than that of FL-LLM, 
mainly due to the fact that the encryption and decryption of FP16 introduce some truncation errors. Second, the accuracy of Method2 is significantly higher than that of other methods for the five datasets, especially on the three datasets of CHIP-CTC, KUAKE-IR, and KUAKE-QIC, which indicates that the split fine-tuning strategy of Method2 can effectively improve the accuracy of the downstream tasks. However, the number of fine-tuned parameters in Method2 is 
about 5 times as many as that of Method1. Table \ref{table6} shows the training and inference time for different methods, where the training time is the average forward and backward computation time for one piece of data from the client. We can conclude that SWMT has the longest training and inference time mainly since the TEE has only CPU processors. For Method1, it shows that the encryption-state operations and data transmission between the TEE and GPU incur a large time overhead. Since only the last $m$ layers of operations in the TEE, Method2 has a shorter training time. Though FL-LLM has a small latency, it cannot guarantee the security of the model parameters and the data.
\vspace{-0.2cm}
\renewcommand{\tabcolsep}{2.0pt}
    \begin{table}[htbp]
        \footnotesize
    	\begin{center}
    			{
    				\renewcommand{\arraystretch}{1.10}
    			\begin{tabular}{c|c|c|c}
    			\bottomrule[0.8pt]
    			& &Training time (s) &Inference time (s) \\ \hline
                    \multirow{2}*{Method1}	&LoRA	&17.06	&2.43	 \\
                    &P-Tuning v2	&11.76 &1.96	 \\ \hline
                    Method2	&4 layers in TEE (25\%,50\%)	&4.33	&0.73 \\ \hline
                    \multirow{2}*{FL-LLM}	&LoRA	&0.48	&0.11	\\
                    &P-Tuning v2 	&0.36	&0.10	\\ \hline
                    \multirow{2}*{SWMT}	&LoRA	&17.92	&3.14	\\
                    &P-Tuning v2 	&17.60	&3.03	\\ \toprule[0.8pt]
    			\end{tabular}
    		}
      \vspace{-0.3cm}
    		\caption{Comparison of average training and inference time for a piece of data in different methods.}\label{table6}
    	\end{center}
     \vspace{-0.5cm}
    \end{table}
    \vspace{-0.1cm}
    
\subsection{ Ablation Experiment for Method2}
\vspace{-0.1cm}
Table \ref{table2} demonstrates the model accuracy and the number of parameters for Method2, segmented by different layers and the selected ratios. As we can see, the more layers deployed in the server or the higher percentage selected, the larger the number of parameters that need to be fine-tuned, which in turn increases the training and inference latency, but brings some performance improvement. In order to balance the accuracy, time cost, and number of parameters, we choose to deploy 4 layers on the server side and the (25\%, 50\%) ratio of parameters to be fine-tuned for comparative experiments in Tab. 1.
\section{CONCLUSION}
\label{sec:majhead}
\vspace{-0.1cm}
In this paper, we propose a secure distributed training framework for LLM based on model slicing to prevent model parameters and data from being maliciously stolen by the server and client. We use model slicing and combine it with the TEE and lightweight encryption strategies to achieve both security and almost lossless model accuracy. We also propose a split fine-tuning scheme to achieve the goals of fast, safety, and high precision. Users can choose between consumer-grade small memory TEE (Method1) and industrial-grade large memory TEE solutions (Method2) in combination with cost consideration.

\vfill\pagebreak

\bibliographystyle{IEEEbib}
\bibliography{strings,refs}

\begin{thebibliography}{10}

\bibitem{peft}
Sourab Mangrulkar, Sylvain Gugger, Lysandre Debut, Younes Belkada, and Sayak Paul,
\newblock ``Peft: State-of-the-art parameter-efficient fine-tuning methods,'' \url{https://github.com/huggingface/peft}, 2022.

\bibitem{hu2022lora}
Edward~J Hu, yelong shen, Phillip Wallis, Zeyuan Allen-Zhu, Yuanzhi Li, Shean Wang, Lu~Wang, and Weizhu Chen,
\newblock ``Lo{RA}: Low-rank adaptation of large language models,''
\newblock in {\em International Conference on Learning Representations}, 2022.

\bibitem{liu2021p}
Xiao Liu, Kaixuan Ji, Yicheng Fu, Weng~Lam Tam, Zhengxiao Du, Zhilin Yang, and Jie Tang,
\newblock ``P-tuning v2: Prompt tuning can be comparable to fine-tuning universally across scales and tasks,''
\newblock {\em arXiv preprint arXiv:2110.07602}, 2021.

\bibitem{kuang2023federatedscopellm}
Weirui Kuang, Bingchen Qian, Zitao Li, Daoyuan Chen, Dawei Gao, Xuchen Pan, Yuexiang Xie, Yaliang Li, Bolin Ding, and Jingren Zhou,
\newblock ``Federatedscope-llm: A comprehensive package for fine-tuning large language models in federated learning,'' 2023.

\bibitem{zhang2023towards}
Jianyi Zhang, Saeed Vahidian, Martin Kuo, Chunyuan Li, Ruiyi Zhang, Guoyin Wang, and Yiran Chen,
\newblock ``Towards building the federated gpt: Federated instruction tuning,''
\newblock {\em arXiv preprint arXiv:2305.05644}, 2023.

\bibitem{mothukuri2021survey}
Viraaji Mothukuri, Reza~M Parizi, Seyedamin Pouriyeh, Yan Huang, Ali Dehghantanha, and Gautam Srivastava,
\newblock ``A survey on security and privacy of federated learning,''
\newblock {\em Future Generation Computer Systems}, vol. 115, pp. 619--640, 2021.

\bibitem{wang2019beyond}
Zhibo Wang, Mengkai Song, Zhifei Zhang, Yang Song, Qian Wang, and Hairong Qi,
\newblock ``Beyond inferring class representatives: User-level privacy leakage from federated learning,''
\newblock in {\em IEEE INFOCOM 2019-IEEE conference on computer communications}. IEEE, 2019, pp. 2512--2520.

\bibitem{8241854}
Le~Trieu Phong, Yoshinori Aono, Takuya Hayashi, Lihua Wang, and Shiho Moriai,
\newblock ``Privacy-preserving deep learning via additively homomorphic encryption,''
\newblock {\em IEEE Transactions on Information Forensics and Security}, vol. 13, no. 5, pp. 1333--1345, 2018.

\bibitem{tramer2018slalom}
Florian Tramer and Dan Boneh,
\newblock ``Slalom: Fast, verifiable and private execution of neural networks in trusted hardware,''
\newblock {\em arXiv preprint arXiv:1806.03287}, 2018.

\bibitem{mckeen2016intel}
Frank McKeen, Ilya Alexandrovich, Ittai Anati, Dror Caspi, Simon Johnson, Rebekah Leslie-Hurd, and Carlos Rozas,
\newblock ``Intel{\textregistered} software guard extensions (intel{\textregistered} sgx) support for dynamic memory management inside an enclave,''
\newblock in {\em Proceedings of the Hardware and Architectural Support for Security and Privacy 2016}, pp. 1--9. 2016.

\bibitem{tdx}
``Intel. intel trust domain extensions,'' \url{https://www.intel.com/content/www/us/en/developer/to ols/trust-domain-extensions/documentation.html}.

\bibitem{agarwal2018cpsgd}
Naman Agarwal, Ananda~Theertha Suresh, Felix Xinnan~X Yu, Sanjiv Kumar, and Brendan McMahan,
\newblock ``cpsgd: Communication-efficient and differentially-private distributed sgd,''
\newblock {\em Advances in Neural Information Processing Systems}, vol. 31, 2018.

\bibitem{wagh2019securenn}
Sameer Wagh, Divya Gupta, and Nishanth Chandran,
\newblock ``Securenn: 3-party secure computation for neural network training.,''
\newblock {\em Proc. Priv. Enhancing Technol.}, vol. 2019, no. 3, pp. 26--49, 2019.

\bibitem{kaplan2016amd}
David Kaplan, Jeremy Powell, and Tom Woller,
\newblock ``Amd memory encryption,''
\newblock {\em White paper}, p.~13, 2016.

\bibitem{alves2004trustzone}
Tiago Alves,
\newblock ``Trustzone: Integrated hardware and software security,''
\newblock {\em Information Quarterly}, vol. 3, pp. 18--24, 2004.

\bibitem{hou2021model}
Jiahui Hou, Huiqi Liu, Yunxin Liu, Yu~Wang, Peng-Jun Wan, and Xiang-Yang Li,
\newblock ``Model protection: Real-time privacy-preserving inference service for model privacy at the edge,''
\newblock {\em IEEE Transactions on Dependable and Secure Computing}, vol. 19, no. 6, pp. 4270--4284, 2021.

\bibitem{shen2022model}
Yun Shen, Xinlei He, Yufei Han, and Yang Zhang,
\newblock ``Model stealing attacks against inductive graph neural networks,''
\newblock in {\em 2022 IEEE Symposium on Security and Privacy (SP)}. IEEE, 2022, pp. 1175--1192.

\bibitem{zhang2021cblue}
Ningyu Zhang, Mosha Chen, Zhen Bi, Xiaozhuan Liang, Lei Li, Xin Shang, Kangping Yin, Chuanqi Tan, Jian Xu, Fei Huang, et~al.,
\newblock ``Cblue: A chinese biomedical language understanding evaluation benchmark,''
\newblock {\em arXiv preprint arXiv:2106.08087}, 2021.

\bibitem{cblue}
``Promptcblue,'' \url{https://github.com/michael-wzhu/PromptCBLUE}, 2023.

\bibitem{zeng2023glm-130b}
Aohan Zeng, Xiao Liu, Zhengxiao Du, Zihan Wang, Hanyu Lai, Ming Ding, Zhuoyi Yang, Yifan Xu, Wendi Zheng, Xiao Xia, Weng~Lam Tam, Zixuan Ma, Yufei Xue, Jidong Zhai, Wenguang Chen, Zhiyuan Liu, Peng Zhang, Yuxiao Dong, and Jie Tang,
\newblock ``{GLM}-130b: An open bilingual pre-trained model,''
\newblock in {\em The Eleventh International Conference on Learning Representations (ICLR)}, 2023.

\end{thebibliography}

\end{document}